\documentclass[11pt,a4paper]{article}
\usepackage[hyperref]{acl2021}
\usepackage{times}
\usepackage{latexsym}

\usepackage{amsfonts}
\usepackage{microtype}
\usepackage{amsmath}
\usepackage{breqn}
\usepackage{amsmath}
\usepackage{amsfonts}
\usepackage{float}
\usepackage{multirow}
\usepackage{graphicx}
\usepackage{subfigure}
\usepackage{caption}
\usepackage{mathrsfs}
\usepackage{booktabs}
\usepackage{listings}
\usepackage{longtable}
\usepackage{float}
\usepackage{algorithm}  
\usepackage{algorithmicx}  
\usepackage{algpseudocode}  
\usepackage{amsmath}  

\aclfinalcopy 

\title{Whitening Sentence Representations for Better \\ Semantics and Faster Retrieval}

\author{\\
Jianlin Su\textsuperscript{\rm 1}, 
Jiarun Cao\textsuperscript{\rm 1}, 
Weijie Liu\textsuperscript{\rm 2}, 
Yangyiwen Ou\textsuperscript{\rm 1},\\
\textsuperscript{\rm 1}Shenzhen Zhuiyi Technology Co., Ltd. \\
\textsuperscript{\rm 2}Tencent Research \\
\small{\{bojonesu, jrcao, owenou\}@wezhuiyi.com, jagerliu@tencent.com}
}

\date{}

\begin{document}
\maketitle
\begin{abstract}

Pre-training models such as BERT have achieved great success in many natural language processing tasks. However, how to obtain better sentence representation through these pre-training models is still worthy to exploit. Previous work has shown that the anisotropy problem is an critical bottleneck for BERT-based sentence representation which hinders the model to fully utilize the underlying semantic features. Therefore, some attempts of boosting the isotropy of sentence distribution, such as flow-based model, have been applied to sentence representations and achieved some improvement. In this paper, we find that the whitening operation in traditional machine learning can similarly enhance the isotropy of sentence representations and achieve competitive results. Furthermore, the whitening technique is also capable of reducing the dimensionality of the sentence representation. Our experimental results show that it can not only achieve promising performance but also significantly reduce the storage cost and accelerate the model retrieval speed.\footnote{The source code of this paper is available at \url{https://github.com/bojone/BERT-whitening}.}

\end{abstract}

\section{Introduction}

The application of deep neural language models~\citep{devlin2018bert,peters2018deep,radford2019language,brown2020language} gained great success in recent years, since they create contextualized word representations that are sensitive to the surrounding context. This trend also stimulates the advance of generating semantic representations of longer piece of text, such as sentences and paragraphs~\citep{arora2016simple}. However, sentence embeddings have been proven to poorly capture the underlying semantics of sentences~\citep{li2020sentence} as the previous work~\citep{gao2019representation,ethayarajh2019contextual,li2020sentence} suggested that the word representations of all words are not isotropic: they are not uniformly distributed with respect to direction. Instead, they occupy a narrow cone in the vector space, and are therefore \textit{anisotropic}. ~\citep{ethayarajh2019contextual} has proved that the contextual word embeddings from the pre-trained model is so anisotropic that any two word embeddings have, on average, a cosine similarity of 0.99. Further investigation from ~\citep{li2020sentence} found that the BERT sentence embedding space suffers from two problems, that is, word frequency biases the embedding space and low-frequency words disperse sparsely, which lead to cause the difficulty of using BERT sentence embedding directly through simple similarity metrics such as dot product or cosine similarity.

To address the problem aforementioned, ~\citep{ethayarajh2019contextual} elaborates on the theoretical reason that leads to the anisotropy problem, as observed in pre-trained models. ~\citep{gao2019representation} designs a novel way to mitigate the degeneration problem by regularizing the word embedding matrix. A recent attempt named BERT-flow~\citep{li2020sentence}, proposed to transform the BERT sentence embedding distribution into a smooth and isotropic Gaussian distribution through normalizing flow~\citep{dinh2014nice}, which is an invertible function parameterized by neural networks. 

Instead of designing a sophisticated method as the previous attempts did, in this paper, we find that a simple and effective post-processing technique -- \textit{whitening} -- is capable enough of tackling the anisotropic problem of sentence embeddings~\citep{reimers2019sentence}. Specifically, we transform the mean value of the sentence vectors to 0 and the covariance matrix to the identity matrix. In addition, we also introduce a dimensionality reduction strategy to facilitate the whitening operation for further improvement the effect of our approach.

The experimental results on 7 standard semantic textual similarity benchmark datasets show that our method can generally improve the model performance and achieve the state-of-the-art results on most of datasets. Meanwhile, by adding the dimensionality reduction operation, our approach can further boost the model performance, as well as naturally optimize the memory storage and accelerate the retrieval speed. 

The main contributions of this paper are summarized as follows:
\begin{itemize}
\item We explore the reason for the poor performance of BERT-based sentence embedding in similarity matching tasks, i.e., it is not in a standard orthogonal basis.
\item A whitening post-processing method is proposed to transform the BERT-based sentence to a standard orthogonal basis while reducing its size.
\item Experimental results on seven semantic textual similarity tasks demonstrate that our method can not only improve model performance significantly, but also reduce vector size.
\end{itemize}

\section{Related Work}

Early attempts on tackling the anisotropic problem have appeared in specific NLP contexts. ~\citep{arora2016simple} first computed the sentence representation for the entire semantic textual similarity dataset, then extracted the top direction from those sentence representations and ﬁnally projected the sentence representation away from it. By doing so, the top direction will inherently encode the common information across the entire dataset. ~\citep{mu2017all} proposed a postprocessing operation is on dense low-dimensional representations with both positive and negative entries, they eliminate the common mean vector and a few top dominating directions from the word vectors, so that renders off-the-shelf representations even stronger. ~\citep{gao2019representation} proposed a novel regularization method to address the anisotropic problem in training natural language generation models. They design a novel way to mitigate the degeneration problem by regularizing the word embedding matrix. As observe that the word embeddings are restricted into a narrow cone, the proposed approach directly increase the size of the aperture of the cone, which can be simply achieved by decreasing the similarity between individual word embeddings.\citep{ethayarajh2019contextual} investigated the inner mechanism of contextual contextualized word representations. They found that upper layers of ELMo, BERT, and GPT-2 produce more context-speciﬁc representations than lower layers. This increased context-speciﬁcity is always accompanied by increased anisotropy. Following up \citep{ethayarajh2019contextual}'s work, \citep{li2020sentence} proposed BERT-flow, in which it transforms the anisotropic sentence embedding distribution to a smooth and isotropic Gaussian distribution through normalizing ﬂows that are learned with an unsupervised objective.

When it comes to state-of-the-art sentence embedding methods, previous work~\citep{conneau2017supervised,cer2017semeval} found that the SNLI datasets are suitable for training sentence embeddings and ~\citep{yang2018learning} proposed a method to train on conversations from Reddit using siamese DAN and siamese transformer networks, which yielded good results on the STS benchmark dataset. ~\citep{cer2018universal} proposed a so-called Universal Sentence Encoder which trains a transformer network and augments unsupervised learning with training on SNLI dataset. In the era of pre-trained methods, ~\citep{humeau2019real} addressed the run-time overhead of the cross-encoder from BERT and presented a method (poly-encoders) to compute a score between context vectors and pre-computed candidate embeddings using attention. ~\citep{reimers2019sentence} is a modiﬁcation of the pretrained BERT network that use siamese and triplet network structures to derive semantically meaningful sentence embeddings that can be compared using cosine-similarity.

\section{Our Approach}
\label{approach}

\subsection{Hypothesis}
\label{Heuristic Hypothesis}
Sentence embedding should be able to intuitively reﬂect the semantic similarity between sentences. When we retrieve semantically similar sentences, we generally encoder the raw sentences into sentence representations, and then calculate the cosine value of their angles for comparison or ranking~\citep{rahutomo2012semantic}. Therefore, a thought-provoking question comes up: what assumptions does cosine similarity make about the input vector? In other words, what preconditions would fit in vectors comparison by cosine similarity? 

We answer this question by studying the geometry of cosine similarity. Geometrically, given two vectors $x \in \mathbb{R}^d$ and $y \in \mathbb{R}^d$, we are aware that inner product of $x$ and $y$ is the product of the Euclidean magnitudes and the cosine of the angle between them. Accordingly, the cosine similarity $\cos(x,y)$ is the inner product of $x$ and $y$ divided by their norms:
\begin{equation}
\label{e1}
    \cos (x,y) = \frac{\sum^d_i x_iy_i}{\sqrt{\sum^d_i x_i^2}\sqrt{\sum^d_i y_i^2}}
\end{equation}

However, the above equation~\ref{e1} is only satisfied when the coordinate basis is Standard Orthogonal Basis. The cosine of the angle has a distinct geometric meaning, but the equation~\ref{e1} is operation-based, which depends on the selected coordinate basis. Therefore, the coordinate formula of the inner product varies with the change of the coordinate basis, and the coordinate formula of the cosine value will also change accordingly.

~\citep{li2020sentence} verified that sentence embedding from BERT~\citep{devlin2018bert} has included sufficient semantics although it is not exploited properly. In this case, if the sentence embeddings perform poorly when equation~\ref{e1} is operated to calculate the cosine value of semantic similarity, the reason may be that the coordinate basis to which the sentence vector belongs is not the Standard Orthogonal Basis. From a statistical point of view, we can infer that it is supposed to ensure each basis vector is independent and uniform when we choose the basis for a set of vectors. If this set of basis is Standard Orthogonal Basis, then the corresponding set of vectors should show isotropy.

To summarize, the above heuristic hypothesis elaborately suggests: if a set of vectors satisfies isotropy, we can assume it is derived from the Standard Orthogonal Basis in which it also indicates that we can calculate the cosine similarity via equation~\ref{e1}. Otherwise, if it is asotropic, we need to transform the original sentence embedding in a way to enforce it being isotropic, and then use the equation~\ref{e1} to calculate the cosine similarity.

\subsection{Whitening Transformation}
\label{Whitening Transformation}
Previous work~\citep{li2020sentence} address the hypothesis in section~\ref{Heuristic Hypothesis} by adopting a flow-based approach. We find that utilizing the whitening operation which is commonly-adopted in machine learning can also achieve comparable gains. 

As far as we are aware that the mean value is 0 and the covariance matrix is a identity matrix with respect to the standard normal distribution. Thus, our goal is to transform the mean value of the sentence vector into 0 and the covariance matrix into the identity matrix. Presumably we have a set of sentence embeddings, which can also be written as a set of row vectors $\{x_i\}^N_{i=1}$, then we carry out a linear transformation in equation~\ref{e2} such that the mean value of $\{x_i\}^N_{i=1}$ is 0 and the covariance matrix is a identity matrix:
\begin{equation}
\label{e2}
    \widetilde x_i = (x_i - \mu )W
\end{equation}
The above equation~\ref{e2} actually corresponds to the whitening operation in machine learning~\citep{christiansen2010data}. In order to let the mean value equals to 0,  we only need to enable:
\begin{equation}
    \mu = \frac{1}{N} \sum_{i=1}^N x_i
\end{equation}

The most difficult part is solving the matrix W. To achieve so, we denote the original covariance matrix of $\{x_i\}^N_{i=1}$ as:

\begin{equation}
    \Sigma = \frac{1}{N} \sum_{i=1}^N (x_i - \mu)^T(x_i - \mu)
\end{equation}

Then we can get the transformed covariance matrix $\widetilde \Sigma$:
\begin{equation}
    \widetilde \Sigma = W^T \Sigma W
\end{equation}
As we specify that the new covariance matrix is an identity matrix, we actually need to solve the equation~\ref{e3} below:

\begin{equation}
\label{e3}   
{
W^{T} \Sigma  W  =  I
}
\end{equation}
Therefore, 

\begin{equation}
\begin{split}
\Sigma &= (W^T)^{-1}W^{-1} \\
       &= (W^{-1})^TW^{-1}
\end{split}
\end{equation}
We are aware that the covariance matrix $\Sigma$ is a positive definite symmetric matrix. The positive definite symmetric matrix satisfies the following form of SVD decomposition~\citep{golub1971singular}:
\begin{equation}
    \Sigma = U \Lambda U^T
\end{equation}
Where $U$ is an orthogonal matrix, $\Lambda$ is a diagonal matrix and the diagonal elements are all positive. Therefore, let $W^{-1} = \sqrt{\Lambda} U^T$, we can obtain the solution:
\begin{equation}
    W = U \sqrt{\Lambda^{-1}}
\end{equation}

\subsection{Dimensionality Reduction}
By far, we already knew that the original covariance matrix of sentence embeddings can be converted into an identity matrix by utilizing the transformation matrix $W = U \sqrt{\Lambda^{-1}}$. Among them, the orthogonal matrix $U$ is a distance-preserving transformation, which means it does not change the relative distribution of the whole data, but transforms the original covariance matrix  $\Sigma$ into the diagonal matrix $\Lambda$.

As far as we know, each diagonal element of the diagonal matrix $\Lambda$ measures the variation of the one-dimensional data in which it is located. If its value is small, it represents that the variation of this dimensional feature is also small and non-significant, even near to a constant. Accordingly, the original sentence vector may only be embedded into a lower dimensional space, and we can remove this dimensional feature while operate dimensionality reduction, where it enables the result of cosine similarity more reasonable and naturally accelerate the speed of vector retrieval as it is directly proportional to the dimensionality.

In fact, the elements in diagonal matrix $\Lambda$ deriving from Singular Value Decomposition~\citep{golub1971singular} has been sorted in the descending order. Therefore, we only need to retain the first $k$ columns of $W$ to achieve this dimensionality reduction effect, which is equivalent to Principal Component Analysis~\citep{abdi2010principal} theoretically. Here, $k$ is an empirical hyperparameter. We refer the 
entire transformation workflow as \texttt{Whitening-$k$}, of which detailed algorithm implementation is shown in Algorithm~\ref{numpy}.

\begin{algorithm}[t]  
  \caption{Whitening-\emph{k} Workflow}  
   {\bf Input:} 
  Existing embeddings $\{x_i\}_{i=1}^N$ and reserved dimensionality $k$
  \begin{algorithmic}[1]  
    \State compute $\mu$ and $\Sigma$ of $\{x_i\}_{i=1}^N$
    \State compute $U, \Lambda, U^{T} = \text{SVD}(\Sigma)$
    \State compute $W = (U\sqrt{\Lambda^{-1}})[:, :k]$
    \For{$i=1,2,\cdots,N$}
        \State $\widetilde{x}_i = (x_i - \mu)W$
    \EndFor
  \end{algorithmic}
  {\bf Output:} Transformed embeddings $\{\widetilde{x}_i\}_{i=1}^N$
  \label{numpy} 
\end{algorithm}

\subsection{Complexity Analysis}
In terms of the computational efficiency on the massive scale of corpora, the mean values $\mu$ and the covariance matrix $\Lambda$ can be calculated recursively. To be more specific, all the above algorithm~\ref{Whitening Transformation} needs are the mean value vector $\mu \in \mathbb{R}^d $ and the covariance matrix $\Sigma \in \mathbb{R}^{d \times d}$(where $d$ is the dimension of word embedding) of the entire sentence vectors $\{x_i\}^N_{i=1}$. Therefore, given the new sentence vector $x_{n+1}$, the mean value can be calculated as:
\begin{equation}
\label{e4}
    \mu_{n+1} = \frac{n}{n+1}\mu_{n}+\frac{1}{n+1}x_{n+1}
\end{equation}
Similarly, convariance matrix is the expectation of $(x_i- \mu)^T(x_i- \mu)$, thus it can be calculated as:
\begin{equation}
\label{e5}
    \Sigma_{n+1} = \frac{n}{n+1}\Sigma_n + \frac{1}{n+1}(x_{n+1}- \mu)^T(x_{n+1}- \mu)
\end{equation}

Therefore, we can conclude that the space complexities of $\mu$ and $\Sigma$ are all $O(1)$ and the time complexities are  $O(N)$, which indicates the effectiveness of our algorithm has reached theoretically optimal. It is reasonable to infer that the algorithm in section~\ref{Whitening Transformation} can obtain the covariance matrix $\Sigma$ and $\mu$ with limited memory storage even in the large-scale corpora. 

\section{Experiment}

To evaluation the effectiveness of the proposed approach, we present our experimental results for various tasks related to semantic textual similarity(STS) tasks under multiple configurations. In the following sections, we first introduce the benchmark datasets in section~\ref{datasets} and our detailed experiment settings in section~\ref{experimental details}. Then, we list our experimental result and in-depth analysis in section~\ref{results}. Furthermore, we evaluate the effect of dimensionality reduction with different settings of dimensionality $k$ in section~\ref{effect}.

\begin{table*}[h]
\centering
\small
\begin{tabular}{lccccccc}
\toprule
& \textbf{STS-B} & \textbf{STS-12} & \textbf{STS-13} & \textbf{STS-14} & \textbf{STS-15} & \textbf{STS-16} & \textbf{SICK-R}\\
\hline\hline
\multicolumn{8}{c}{\textit{Published in \citep{reimers2019sentence}}} \\
\text{Avg. GloVe embeddings} & 58.02 & 55.14 & 70.66 & 59.73 & 68.25 & 63.66 & 53.76 \\
\text{Avg. BERT embeddings} & 46.35 & 38.78 & 57.98 & 57.98 & 63.15 & 61.06 & 58.40 \\
\text{BERT CLS-vector} & 16.50 & 20.16 & 30.01 & 20.09 & 36.88 & 38.03 & 42.63 \\
\hline\hline
\multicolumn{8}{c}{\textit{Published in \citep{li2020sentence}}} \\
$\text{BERT}_{\text{base}}\text{-first-last-avg}$ & 59.04 & 57.84 & 61.95 & 62.48 & 70.95 & 69.81 & 63.75\\
$\text{BERT}_{\text{base}}\text{-flow}\,(\text{NLI})$ & 58.56 & 59.54 & 64.69 & 64.66 & 72.92 & 71.84 & \textbf{65.44}\\
$\text{BERT}_{\text{base}}\text{-flow}\,(\text{target})$ & 70.72 & 63.48 & 72.14 & 68.42 & 73.77 & \textbf{75.37} & 63.11\\
\hline
\multicolumn{8}{c}{\textit{Our implementation}} \\
$\text{BERT}_{\text{base}}\text{-first-last-avg}$ & 59.04 & 57.86 & 61.97 & 62.49 & 70.96 & 69.76 & 63.75\\
$\text{BERT}_{\text{base}}\text{-whitening}\,(\text{NLI})$ & 68.19($\color{green}{\uparrow}$) & 61.69($\color{green}{\uparrow}$) & 65.70($\color{green}{\uparrow}$) & 66.02($\color{green}{\uparrow}$) & \textbf{75.11}($\color{green}{\uparrow}$) & 73.11($\color{green}{\uparrow}$) & 63.6($\color{red}{\downarrow}$)\\
$\text{BERT}_{\text{base}}\text{-whitening-256}\,(\text{NLI})$ & 67.51($\color{green}{\uparrow}$) & 61.46($\color{green}{\uparrow}$) & 66.71($\color{green}{\uparrow}$) & 66.17($\color{green}{\uparrow}$) & 74.82($\color{green}{\uparrow}$) & 72.10($\color{green}{\uparrow}$) & 64.9($\color{red}{\downarrow}$)\\
$\text{BERT}_{\text{base}}\text{-whitening}\,(\text{target})$ & 71.34($\color{green}{\uparrow}$) & 63.62($\color{green}{\uparrow}$) & 73.02($\color{green}{\uparrow}$) & \textbf{69.23}($\color{green}{\uparrow}$) & 74.52($\color{green}{\uparrow}$) & 72.15($\color{red}{\downarrow}$) & 60.6($\color{red}{\downarrow}$)\\
$\text{BERT}_{\text{base}}\text{-whitening-256}\,(\text{target})$ & \textbf{71.43}($\color{green}{\uparrow}$) & \textbf{63.89}($\color{green}{\uparrow}$) & \textbf{73.76}($\color{green}{\uparrow}$) & 69.08($\color{green}{\uparrow}$) & 74.59($\color{green}{\uparrow}$) & 74.40($\color{red}{\downarrow}$) & 62.2($\color{red}{\downarrow}$)\\
\hline\hline
\multicolumn{8}{c}{\textit{Published in \citep{li2020sentence}}} \\
$\text{BERT}_{\text{large}}\text{-first-last-avg}$ & 59.56 & 57.68 & 61.37 & 61.02 & 68.04 & 70.32 & 60.22\\
$\text{BERT}_{\text{large}}\text{-flow}\,(\text{NLI})$ & 68.09 & 61.72 & 66.05 & 66.34 & 74.87 & 74.47 & \textbf{64.62}\\
$\text{BERT}_{\text{large}}\text{-flow}\,(\text{target})$ & 72.26 & \textbf{65.20} & 73.39 & 69.42 & 74.92 & \textbf{77.63} & 62.50\\
\hline
\multicolumn{8}{c}{\textit{Our implementation}} \\
$\text{BERT}_{\text{large}}\text{-first-last-avg}$ & 59.59 & 57.73 & 61.17 & 61.18 & 68.07 & 70.25 & 60.34\\
$\text{BERT}_{\text{large}}\text{-whitening}\,(\text{NLI})$ & 68.54($\color{green}{\uparrow}$) & 62.54($\color{green}{\uparrow}$) & 67.31($\color{green}{\uparrow}$) & 67.12($\color{green}{\uparrow}$) & 75.00($\color{green}{\uparrow}$) & 76.29($\color{green}{\uparrow}$) & 62.4($\color{red}{\downarrow}$)\\
$\text{BERT}_{\text{large}}\text{-whitening-384}\,(\text{NLI})$ & 68.60($\color{green}{\uparrow}$) & 62.28($\color{green}{\uparrow}$) & 67.88($\color{green}{\uparrow}$) & 67.01($\color{green}{\uparrow}$) & \textbf{75.49}($\color{green}{\uparrow}$) & 75.46($\color{green}{\uparrow}$) & 63.8($\color{red}{\downarrow}$)\\
$\text{BERT}_{\text{large}}\text{-whitening}\,(\text{target})$ & 72.14($\color{red}{\downarrow}$) & 64.02($\color{red}{\downarrow}$) & 72.67($\color{red}{\downarrow}$) & 68.93($\color{red}{\downarrow}$) & 73.57($\color{red}{\downarrow}$) & 72.52($\color{red}{\downarrow}$) & 59.3($\color{red}{\downarrow}$)\\
$\text{BERT}_{\text{large}}\text{-whitening-384}\,(\text{target})$ & \textbf{72.48}($\color{green}{\uparrow}$) & 64.34($\color{red}{\downarrow}$) & \textbf{74.60}($\color{green}{\uparrow}$) & \textbf{69.64}($\color{green}{\uparrow}$) & 74.68($\color{red}{\downarrow}$) & 75.90($\color{red}{\downarrow}$) & 60.8($\color{red}{\downarrow}$)\\
\bottomrule
\end{tabular}
\small
\caption{Results without supervision of NLI. We report the Spearman’s rank correlation score as $\rho \times 100$ between the cosine similarity of sentence embeddings and the gold labels on multiple datasets. $\color{green}{\uparrow}$ denotes outperformance over its BERT-flow baseline and $\color{red}{\downarrow}$ denotes underperformance.}
\label{table1}

\end{table*}

\begin{table*}[h]

\centering
\small
\begin{tabular}{lccccccc}
\toprule
& \textbf{STS-B} & \textbf{STS-12} & \textbf{STS-13} & \textbf{STS-14} & \textbf{STS-15} & \textbf{STS-16} & \textbf{SICK-R}\\
\hline\hline
\multicolumn{8}{c}{\textit{Published in \citep{reimers2019sentence}}} \\
$\text{InferSent - Glove}$ & 68.03 & 52.86 & 66.75 & 62.15 & 72.77 & 66.86 & 65.65 \\
$\text{USE}$ & 74.92 & 64.49 & 67.80 & 64.61 & 76.83 & 73.18 & 76.69 \\
$\text{SBERT}_{\text{base}}\text{-NLI}$ & 77.03 & 70.97 & 76.53 & 73.19 & 79.09 & 74.30 & 72.91 \\
$\text{SBERT}_{\text{large}}\text{-NLI}$ & 79.23 & 72.27 & 78.46 & 74.90 & 80.99 & 76.25 & 73.75 \\
$\text{SRoBERTa}_{\text{base}}\text{-NLI}$ & 77.77 & \textbf{71.54} & 72.49 & 70.80 & 78.74 & 73.69 & 74.46 \\
$\text{SRoBERTa}_{\text{large}}\text{-NLI}$ & 79.10 & \textbf{74.53} & 77.00 & 73.18 & 81.85 & 76.82 & 74.29 \\
\hline\hline
\multicolumn{8}{c}{\textit{Published in \citep{li2020sentence}}} \\
$\text{SBERT}_{\text{base}}\text{-NLI-first-last-avg}$ & 78.03 & 68.37 & 72.44 & 73.98 & 79.15 & 75.39 & 74.07\\
$\text{SBERT}_{\text{base}}\text{-NLI-flow}\,(\text{NLI})$ & 79.10 & 67.75 & 76.73 & 75.53 & 80.63 & 77.58 & \textbf{78.03} \\
$\text{SBERT}_{\text{base}}\text{-NLI-flow}\,(\text{target})$ & \textbf{81.03} & 68.95 & 78.48 & 77.62 & 81.95 & 78.94 & 74.97 \\
\hline
\multicolumn{8}{c}{\textit{Our implementation}} \\
$\text{SBERT}_{\text{base}}\text{-NLI-first-last-avg}$ & 77.63 & 68.70 & 74.37 & 74.73 & 79.65 & 75.21 & 74.84\\
$\text{SBERT}_{\text{base}}\text{-NLI-whitening}\,(\text{NLI})$ & 78.66($\color{red}{\downarrow}$) & 69.11($\color{green}{\uparrow}$) & 75.79($\color{red}{\downarrow}$) & 75.76($\color{green}{\uparrow}$) & 82.31($\color{green}{\uparrow}$) & \textbf{79.61}($\color{green}{\uparrow}$) & 76.33($\color{red}{\downarrow}$)\\
$\text{SBERT}_{\text{base}}\text{-NLI-whitening-256}\,(\text{NLI})$ & 79.16($\color{green}{\uparrow}$) & 69.87($\color{green}{\uparrow}$) & 77.11($\color{green}{\uparrow}$) & 76.13($\color{green}{\uparrow}$) & \textbf{82.73}($\color{green}{\uparrow}$) & 78.08($\color{green}{\uparrow}$) & 76.44($\color{red}{\downarrow}$)\\
$\text{SBERT}_{\text{base}}\text{-NLI-whitening}\,(\text{target})$ & 80.50($\color{red}{\downarrow}$) & 69.01($\color{green}{\uparrow}$) & 78.10($\color{red}{\downarrow}$) & 77.04($\color{red}{\downarrow}$) & 80.83($\color{red}{\downarrow}$) & 77.93($\color{red}{\downarrow}$) & 72.54($\color{red}{\downarrow}$)\\
$\text{SBERT}_{\text{base}}\text{-NLI-whitening-256}\,(\text{target})$ & 80.80($\color{red}{\downarrow}$) & 69.97($\color{green}{\uparrow}$) & \textbf{79.48}($\color{green}{\uparrow}$) & \textbf{78.12}($\color{green}{\uparrow}$) & 81.60($\color{red}{\downarrow}$) & 79.07($\color{green}{\uparrow}$) & 75.06($\color{green}{\uparrow}$)\\
\hline\hline
\multicolumn{8}{c}{\textit{Published in \citep{li2020sentence}}} \\
$\text{SBERT}_{\text{large}}\text{-NLI-first-last-avg}$ & 78.45 & 68.69 & 75.63 & 75.55 & 80.35 & 76.81 & 74.93 \\
$\text{SBERT}_{\text{large}}\text{-NLI-flow}\,(\text{NLI})$ & 79.89 & 69.61 & 79.45 & 77.56 & 82.48 & 79.36 & \textbf{77.73} \\
$\text{SBERT}_{\text{large}}\text{-NLI-flow}\,(\text{target})$ & 81.18 & 70.19 & \textbf{80.27} & 78.85 & 82.97 & 80.57 & 74.52 \\
\hline
\multicolumn{8}{c}{\textit{Our implementation}} \\
$\text{SBERT}_{\text{large}}\text{-NLI-first-last-avg}$ & 79.16 & 70.00 & 76.55 & 76.33 & 80.40 & 77.02 & 76.56\\
$\text{SBERT}_{\text{large}}\text{-NLI-whitening}\,(\text{NLI})$ & 79.55($\color{red}{\downarrow}$) & 70.41($\color{green}{\uparrow}$) & 76.78($\color{red}{\downarrow}$) & 76.88($\color{red}{\downarrow}$) & 82.84($\color{green}{\uparrow}$) & \textbf{81.19}($\color{green}{\uparrow}$) & 75.93($\color{red}{\downarrow}$)\\
$\text{SBERT}_{\text{large}}\text{-NLI-whitening-384}\,(\text{NLI})$ & 80.70($\color{green}{\uparrow}$) & 70.97($\color{green}{\uparrow}$) & 78.36($\color{red}{\downarrow}$) & 77.64($\color{green}{\uparrow}$) & \textbf{83.32}($\color{green}{\uparrow}$) & 80.98($\color{green}{\uparrow}$) & 77.10($\color{red}{\downarrow}$)\\
$\text{SBERT}_{\text{large}}\text{-NLI-whitening}\,(\text{target})$ & 81.10($\color{red}{\downarrow}$) & 69.95($\color{red}{\downarrow}$) & 77.76($\color{red}{\downarrow}$) & 77.56($\color{red}{\downarrow}$) & 80.78($\color{red}{\downarrow}$) & 77.40($\color{red}{\downarrow}$) & 71.69($\color{red}{\downarrow}$)\\
$\text{SBERT}_{\text{large}}\text{-NLI-whitening-384}\,(\text{target})$ & \textbf{82.22}($\color{green}{\uparrow}$) & 71.25($\color{green}{\uparrow}$) & 80.05($\color{red}{\downarrow}$) & \textbf{78.96}($\color{green}{\uparrow}$) & 82.53($\color{red}{\downarrow}$) & 80.36($\color{red}{\downarrow}$) & 74.05($\color{red}{\downarrow}$)\\
\bottomrule
\end{tabular}
\small

\caption{Results with supervision of NLI. We report the Spearman’s rank correlation score as $\rho \times 100$ between the cosine similarity of sentence embeddings and the gold labels on multiple datasets.  $\color{green}{\uparrow}$ denotes outperformance over its SBERT-flow baseline and $\color{red}{\downarrow}$ denotes underperformance.}
\label{table2}
\end{table*}

\subsection{Datasets}
\label{datasets}
We compare the model performance with baselines for STS tasks without any specific training data as~\citep{reimers2019sentence} does. 7 datasets including STS 2012-2016 tasks~\citep{agirre2012semeval,agirre2013sem,agirre2014semeval,agirre2015semeval,agirre2016semeval}, the STS benchmark~\citep{cer2017semeval} and the SICK-Relatedness dataset~\citep{marelli2014sick} are adopted as our benchmarks for evalutation. For each sentence pair, these datasets  provide a standard semantic similarity measurement ranging from 0 to 5. We adopt the Spearman’s rank correlation between the cosine-similarity of the sentence embeddings and the gold labels, since ~\citep{reimers2019sentence} suggested it is the most reasonable metrics in STS tasks. The evaluation procedure is kept as same as ~\citep{li2020sentence}, of which we first encode each raw sentence text into sentence embedding, then calculate the cosine similarities between input sentence embedding pairs as our predicted similarity scores. 
\subsection{Experimental Settings and Baselines}

\label{experimental details}
\paragraph{Baselines. } We compare the performanc with the following baselines. In the unsupervised STS, \texttt{Avg. GloVe embeddings} denotes that we adopt GloVe~\citep{pennington2014glove} as the sentence embedding. Similarly, \texttt{Avg. BERT embeddings} and \texttt{BERT CLS-vector} denotes that we use raw BERT~\citep{devlin2018bert} with and without using the \texttt{CLS}-token output. In the surpervised STS, \texttt{USE} denotes Universal Sentence Encoder~\citep{cer2018universal} which replaces the LSTM with a Transformer. While \texttt{SBERT-NLI} and \texttt{SRoBERTa-NLI} correspond to the BERT and RoBERTa~\citep{liu2019roberta} model trained on a combined NLI dataset (consitutuing SNLI~\citep{bowman2015large} and MNLI~\citep{williams2017broad}) with the Sentence-BERT training approach \citep{reimers2019sentence}.

\paragraph{Experimental details. }Since the BERT-flow(\text{NLI/target}) is the primary baseline we are compared to, we basically align to their experimental settings and symbols. Concretely, we also use both $\texttt{BERT}_{\texttt{base}}$ and $\texttt{BERT}_{\texttt{large}}$ in our experiments. We choose \texttt{-first-last-avg}\footnote{In \citep{li2020sentence}, it is marked as \textit{-last2avg}, but it is actually \texttt{-first-last-avg} in its source code.} as our default configuration as averaging the first and the last layers of BERT can stably achieve better performance compared to only averaging the last one layer. 
Similar to \citep{li2020sentence}, we leverage the full target dataset (including all sentences in train, development, and test sets, and excluding all labels) to calculate the whitening parameters $W$ and $\mu$ through the unsupervised approach as described in Section \ref{Whitening Transformation}. These model are symbolized as \texttt{-whitening(target)}. Furthermore, \texttt{-whitening(NLI)} denotes the whitening parameters are obtained on the NLI corpus. \texttt{-whitening-256(target/NLI)} and \texttt{-whitening-384(target/NLI)} indicates that through our whitening method, the output embedding size is reduced to 256 and 384, respectively.

\subsection{Results}
\label{results}
\paragraph{Without supervision of NLI.}

As shown in Table~\ref{table1}, the raw BERT and GloVe sentence embedding unsuprisingly obtain the worst performance on these datasets. Under the $\texttt{BERT}_{\texttt{base}}$ settings, our approach consistently outperforms the BERT-flow and achieves state-of-the-art results with 256 sentence embedding dimensionality on STS-B, STS-12, STS-13, STS-14, STS-15 datasets respectively. When we switch to $\texttt{BERT}_{\texttt{large}}$, the better results achieved if the dimensionality of sentence embedding set to 384. Our approach still gains the competitive results on most of the datasets compared to BERT-flow, and achieves the state-of-the-art results by roughly 1 point on STS-B, STS-13, STS-14 datasets.

\paragraph{With supervision of NLI.}

In Table~\ref{table2}, the $\texttt{SBERT}_{\texttt{base}}$ and $\texttt{SBERT}_{\texttt{large}}$ are trained on the NLI dataset with supervised labels through the approach in \citep{reimers2019sentence}. It could be observed that our $\texttt{SBERT}_{\texttt{base}}\texttt{-whitening}$ outperforms $\texttt{BERT}_{\texttt{base}}\texttt{-flow}$ on the STS-13, STS-14, STS-15, STS-16 tasks, and $\texttt{SBERT}_{\texttt{large}}\texttt{-whitening}$ obtains better result $\texttt{BERT}_{\texttt{large}}\texttt{-flow}$ on STS-B, STS-14, STS-15, STS-16 tasks. These experimental results show that our whitening method can further improve the performance of SBERT, even though it has been trained under the supervision of the NLI dataset.

\begin{figure*}[!htbp]
  \centering
  \subfigure[$\text{BERT}_{\text{base}}\,(\text{NLI})$]{
    \includegraphics[width=0.22\textwidth]{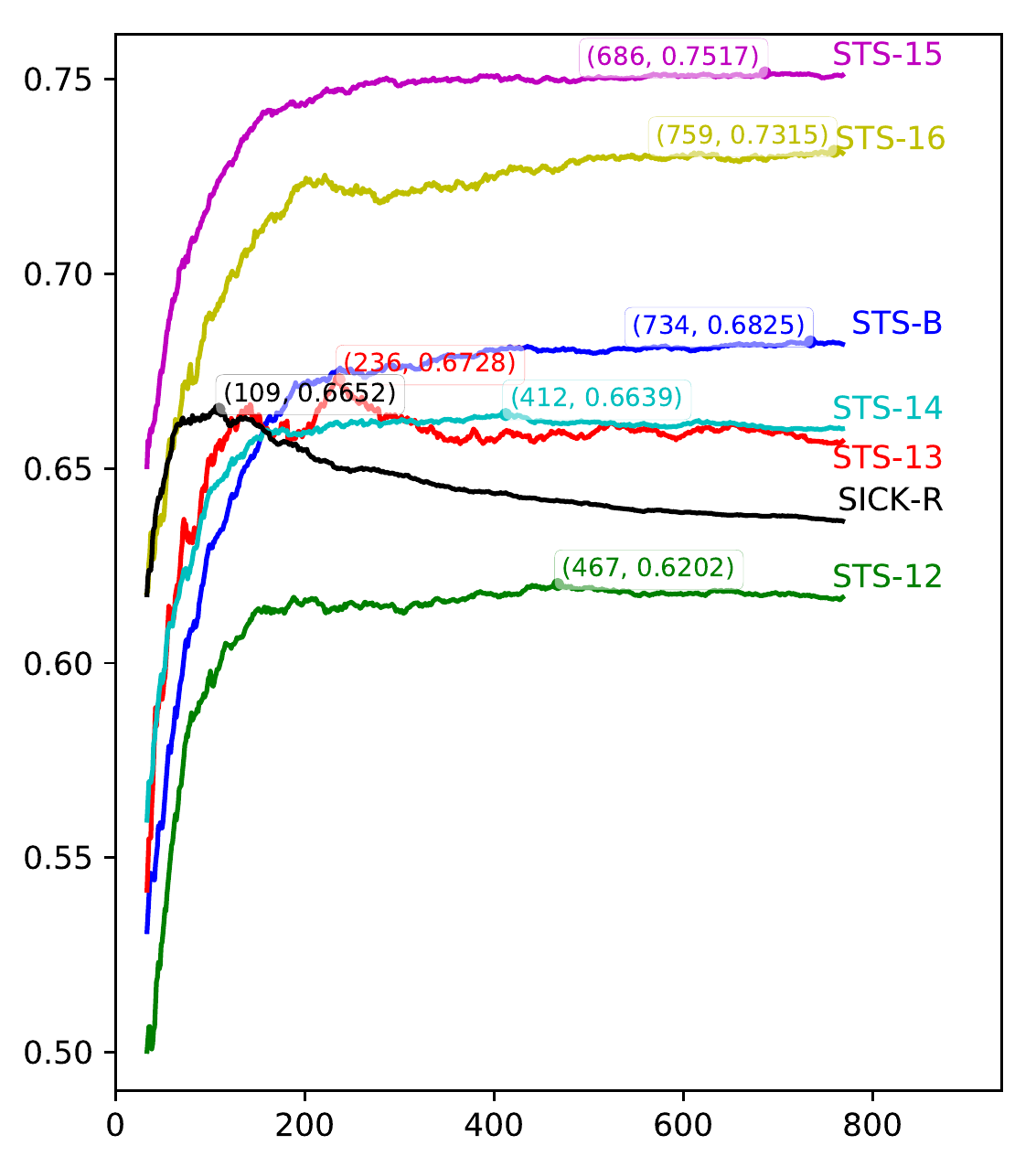}}
  \subfigure[$\text{BERT}_{\text{base}}\,(\text{target})$]{
    \includegraphics[width=0.22\textwidth]{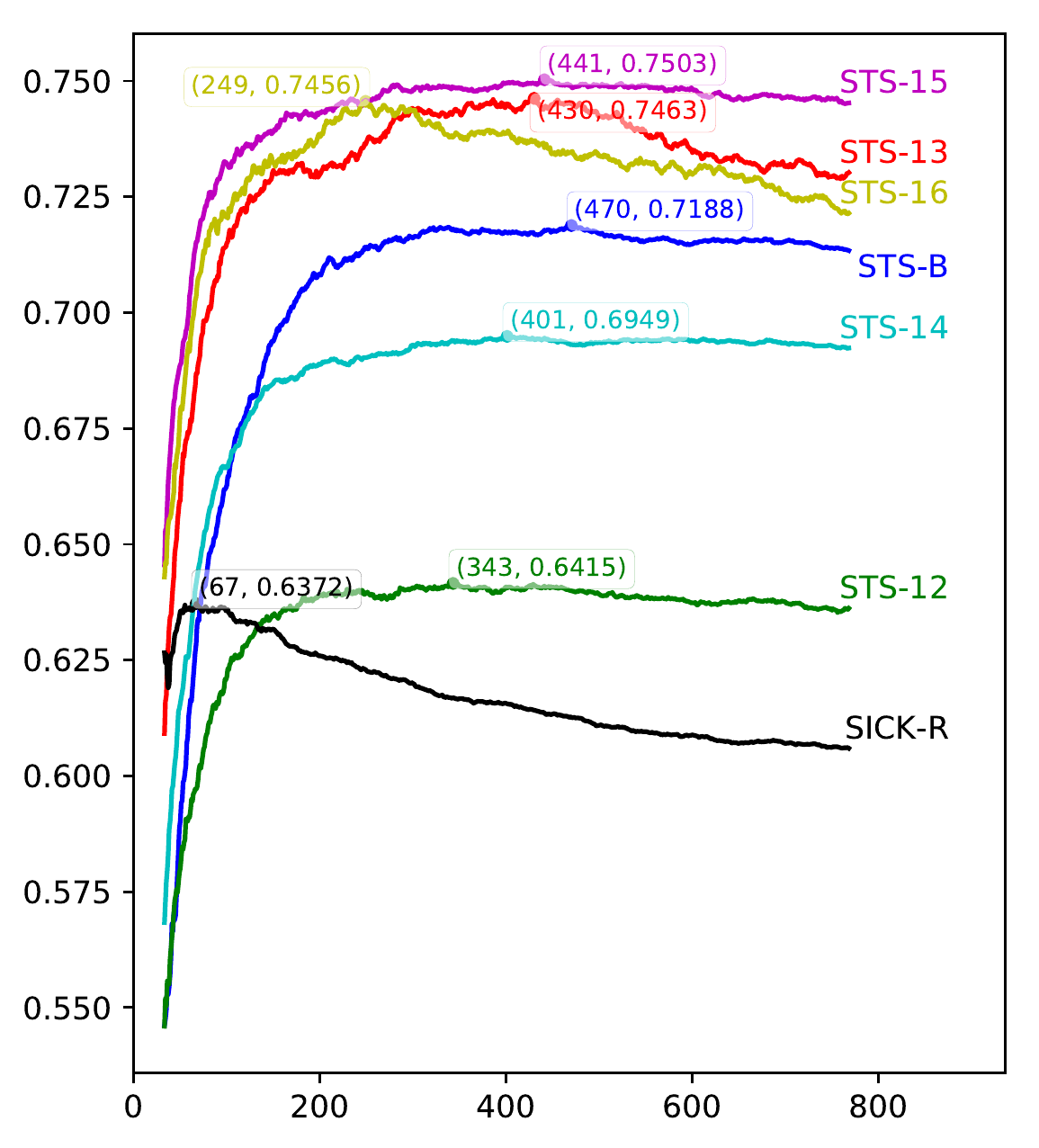}}
  \subfigure[$\text{SBERT}_{\text{base}}\text{-NLI}\,(\text{NLI})$]{
    \includegraphics[width=0.22\textwidth]{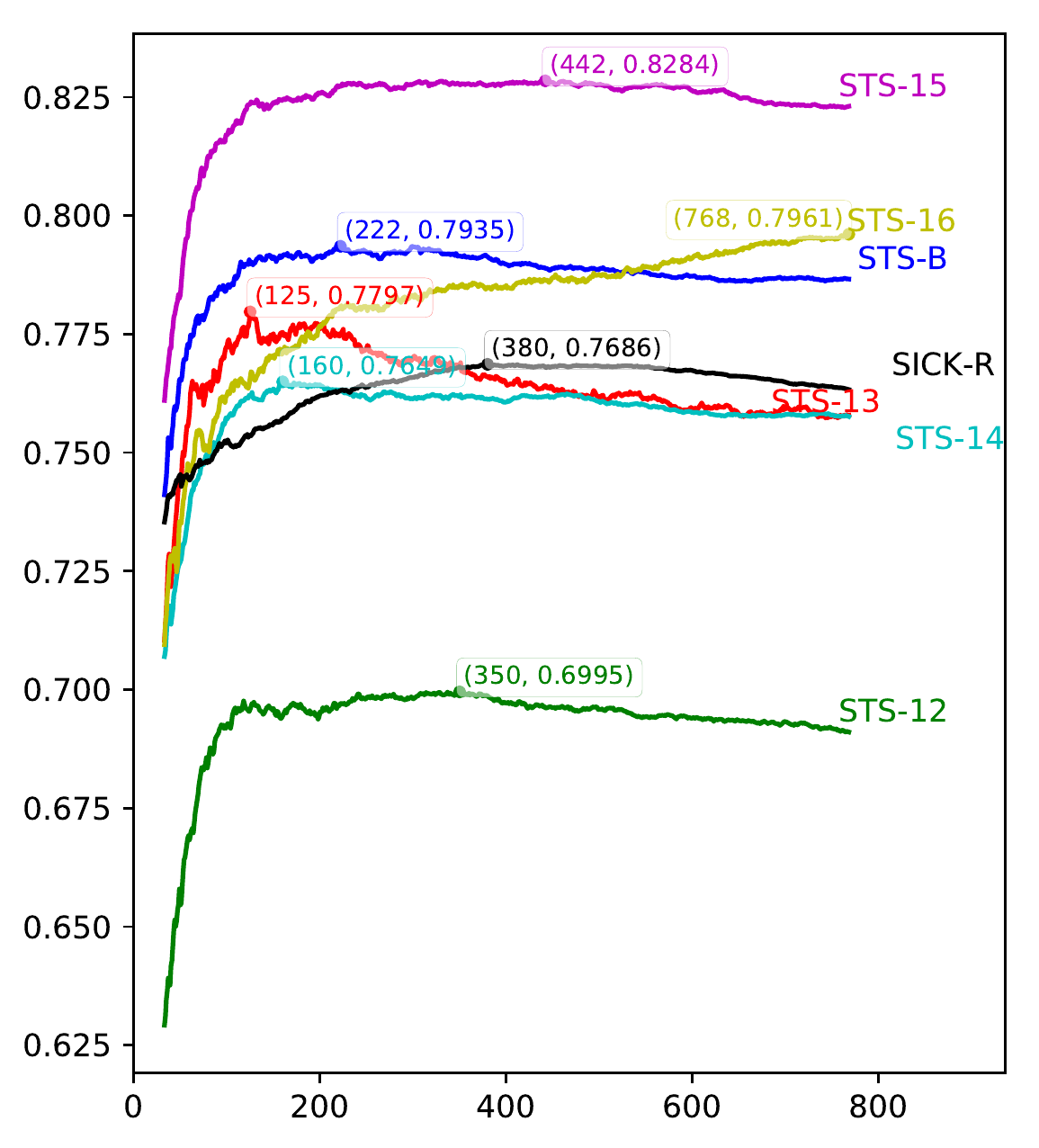}}
  \subfigure[$\text{SBERT}_{\text{base}}\text{-NLI}\,(\text{target})$]{
    \includegraphics[width=0.22\textwidth]{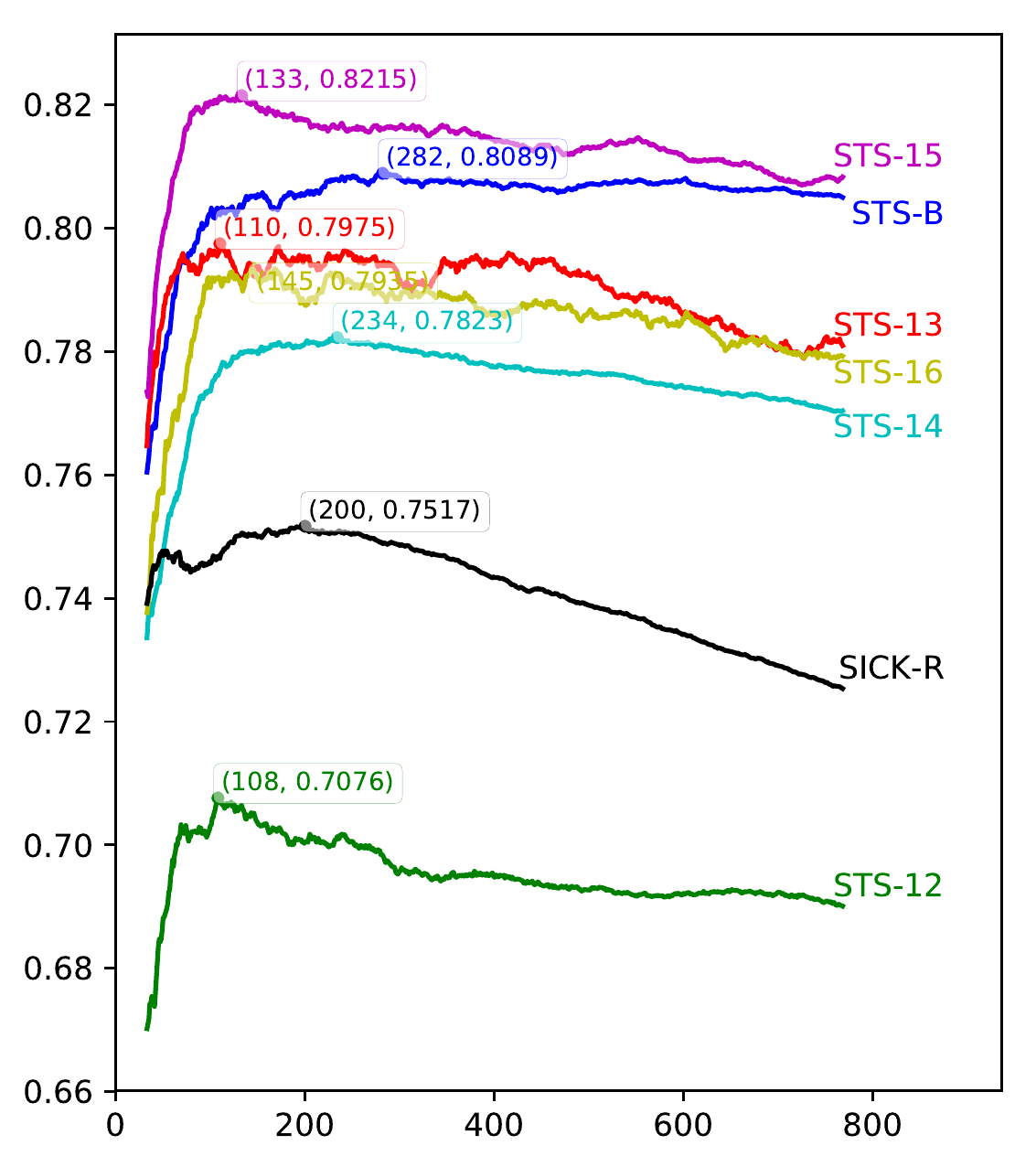}}
  \subfigure[$\text{BERT}_{\text{large}}\,(\text{NLI})$]{
    \includegraphics[width=0.22\textwidth]{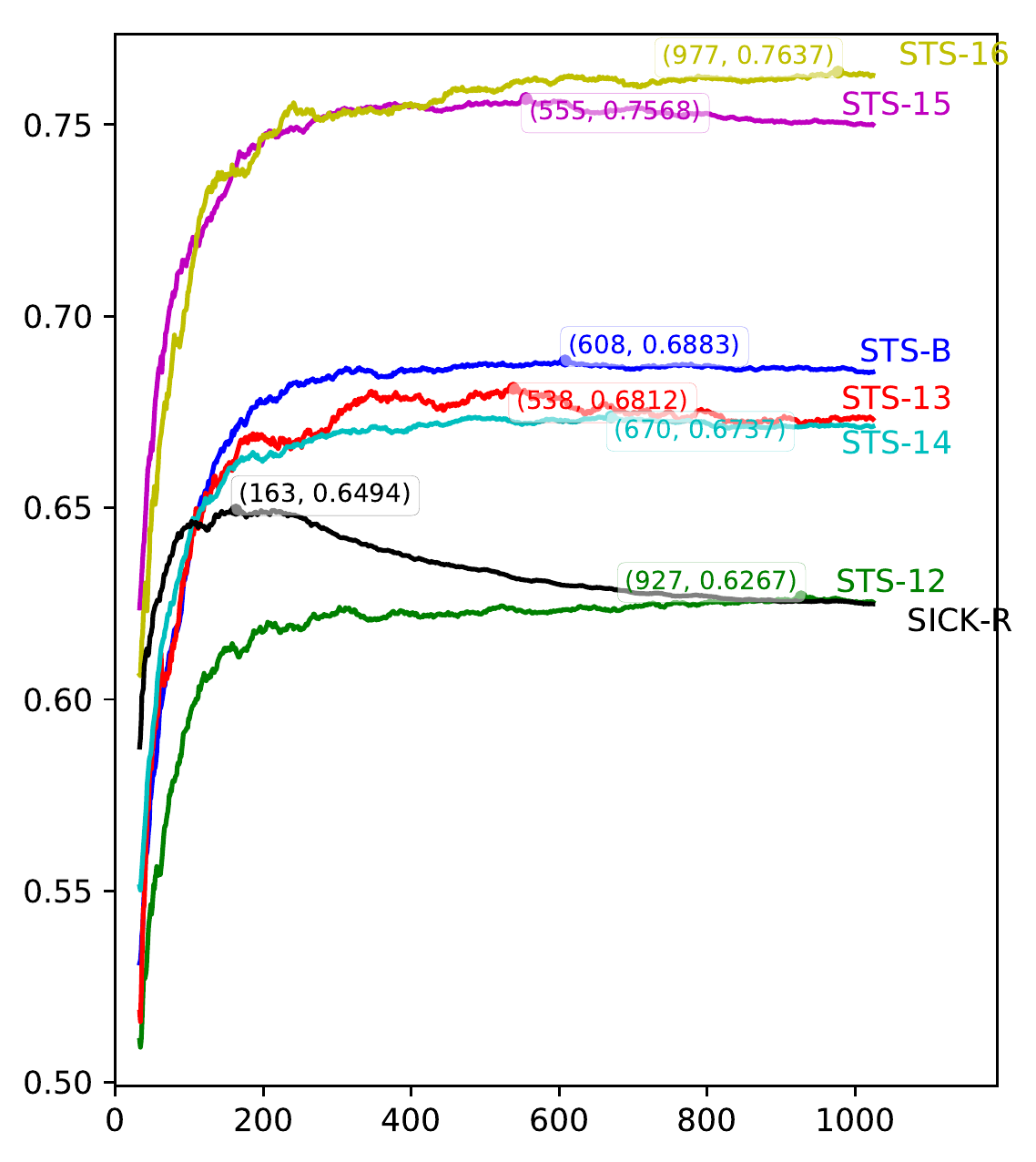}}
  \subfigure[$\text{BERT}_{\text{large}}\,(\text{target})$]{
    \includegraphics[width=0.22\textwidth]{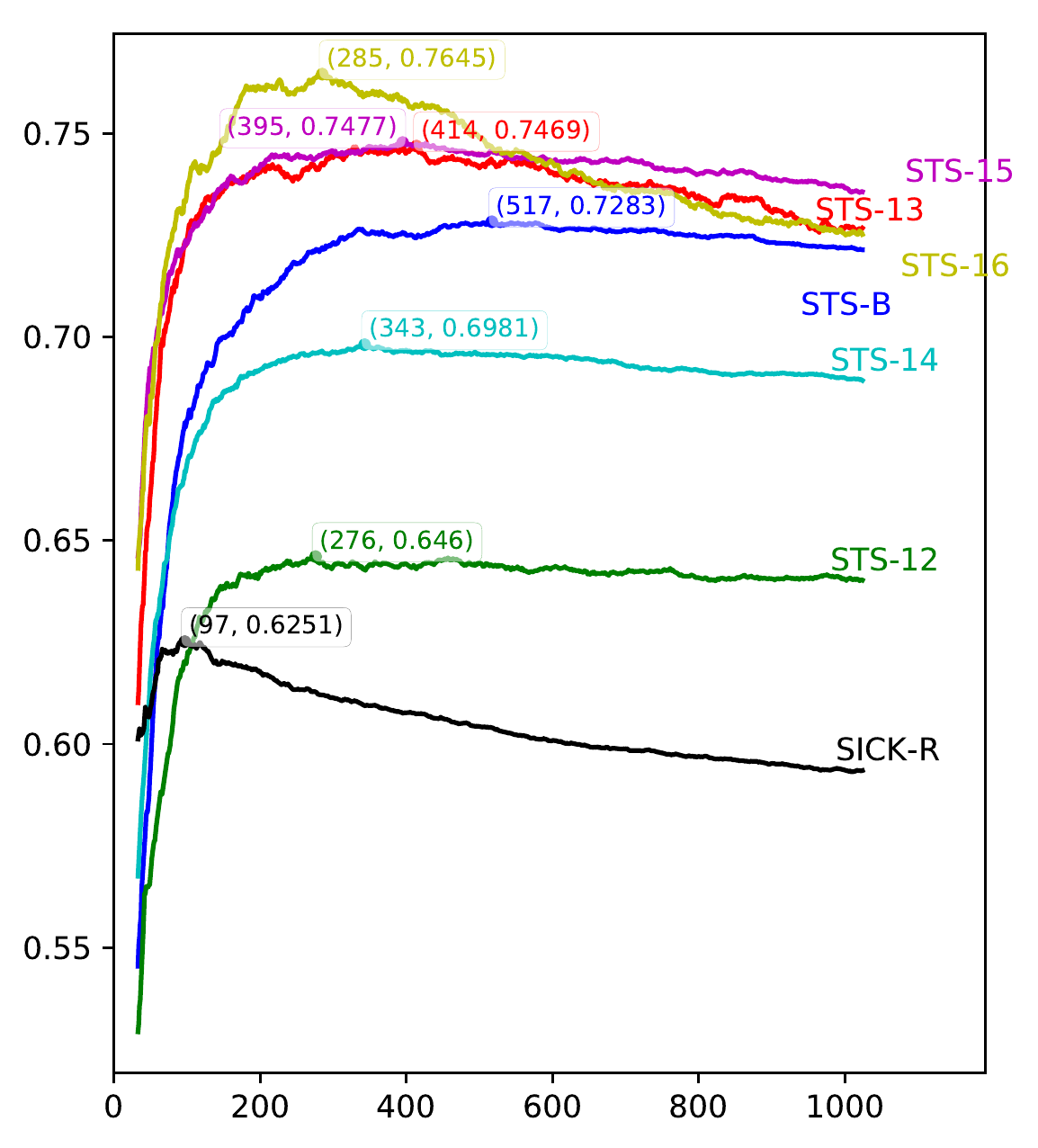}}
  \subfigure[$\text{SBERT}_{\text{large}}\text{-NLI}\,(\text{NLI})$]{
    \includegraphics[width=0.22\textwidth]{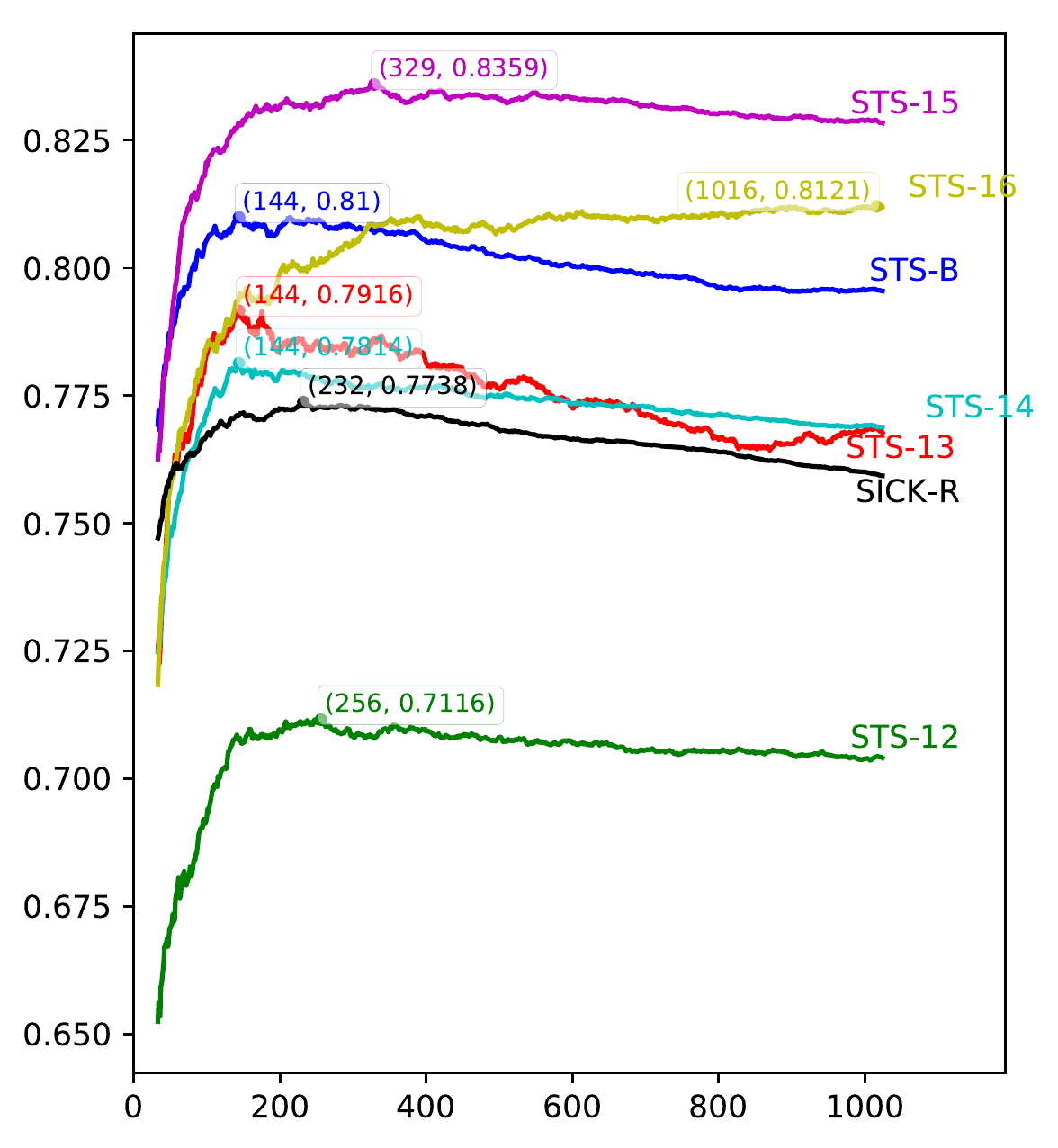}}
  \subfigure[$\text{SBERT}_{\text{large}}\text{-NLI}\,(\text{target})$]{
    \includegraphics[width=0.22\textwidth]{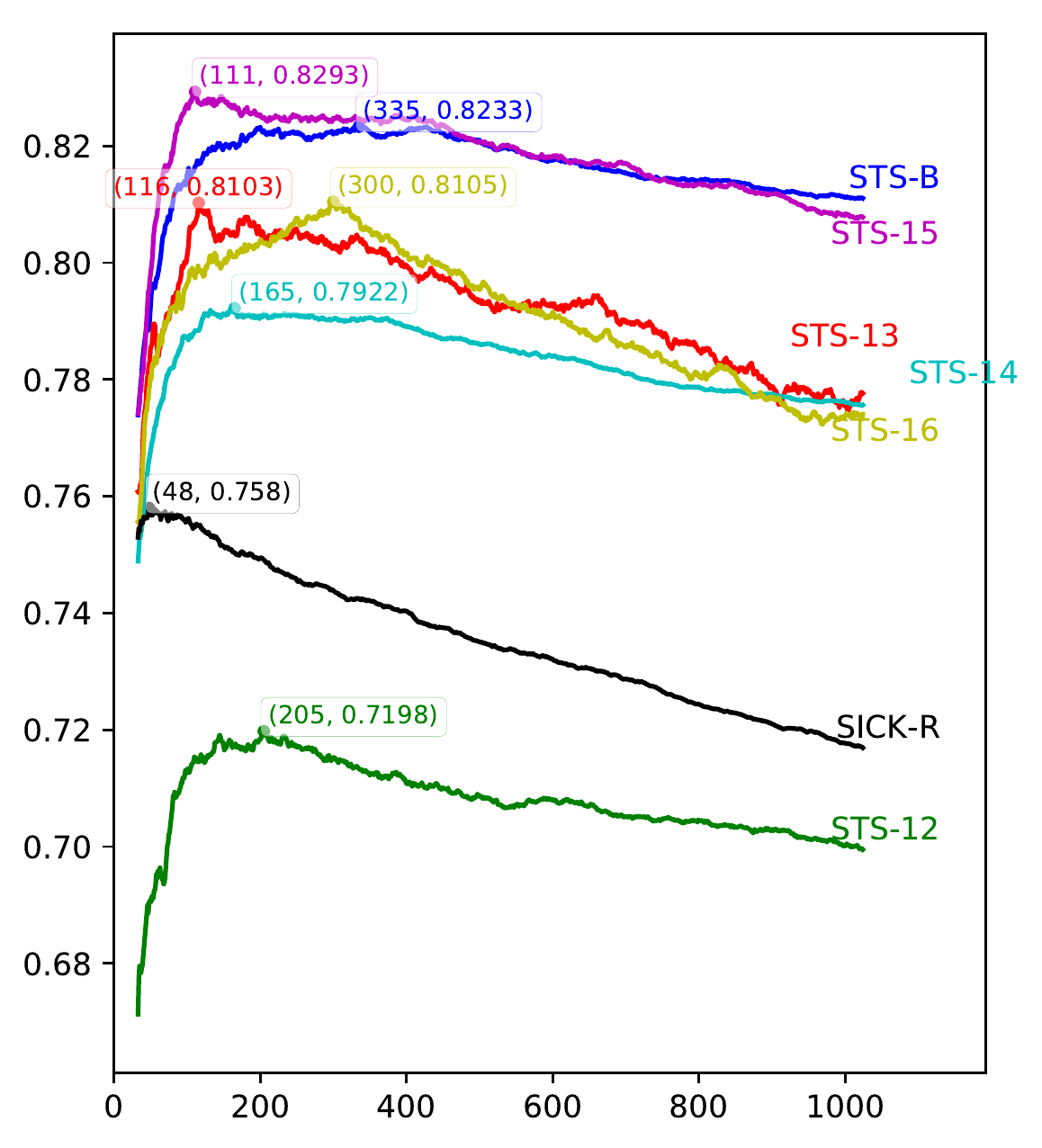}}
  \caption{Effect of different dimensionality $k$ with BERT-whitening on each aforementioned tasks. the $x$ axis is the reserved dimension of sentence embeddings, the $y$ axis is the Spearman's correlation coefficient. The marked point on each sub-figure is the location of the optimal result.}
\label{fig1}

\end{figure*}

\subsection{Effect of Dimensionality $k$}
\label{effect}
Dimensionality reduction is a crucial feature, because reduction of vector size brings about smaller memory occupation and a faster retrieval for downstream vector search engines. The dimensionality $k$ is a hyperparameter of reserved dimension of sentence embeddings, which can affect the model performance by large margin. Therefore, we carry out experiment to test the variation of Spearman's correlation coefficient of the model with the change of dimensionality $k$. Figure~\ref{fig1} presents the variation curve of model performance under $\texttt{BERT}_{\texttt{base}}$ and $\texttt{BERT}_{\texttt{large}}$ embeddings. For most tasks, reducing the dimension of the sentence vector to its one of third is an relatively optimal solution, in which its performance is at the edge of increasing point.

In the SICK-R results in Table~\ref{table1}, although our $\texttt{BERT}_{\texttt{base}}\texttt{-whitening-256}\,(\texttt{NLI})$ is not as effective as $\texttt{BERT}_{\texttt{base}}\texttt{-flow}\,(\texttt{NLI})$, our model has a competitive advantage, i.e., the smaller embedding size (256 vs. 768). Furthermore, as presented in Figure \ref{fig1}(a), the correlation score of our $\texttt{BERT}_{\texttt{base}}\texttt{-whitening}\,(\texttt{NLI})$ raises to 66.52 when the embedding size is set to 109, which outperforms the $\texttt{BERT}_{\texttt{base}}\texttt{-flow}\,(\texttt{NLI})$ by 1.08 point. Besides, other tasks can also achieve better performances by choosing $k$ carefully.

\section{Conclusion}

In this work, we explore an alternative approach to alleviate the anisotropy problem of sentence embedding. Our approach is based on the whitening operation in machine learning, where experimental results indicate our method is simple but effective  on 7 semantic similarity benchmark datasets. Besides, we also find that introduce dimensionality reduction operation can further boost the model performance, and naturally optimize the memory storage and accelerate the retrieval speed.

\bibliographystyle{acl_natbib}
\bibliography{anthology,acl2021}



\end{document}